\documentclass{article}

\usepackage[preprint]{neurips_2021}
\usepackage[utf8]{inputenc} 
\usepackage[T1]{fontenc}    
\usepackage{hyperref}       
\usepackage{url}            
\usepackage{booktabs}       
\usepackage{amsfonts}       
\usepackage{nicefrac}       
\usepackage{microtype}      
\usepackage{xcolor}         
\usepackage{graphicx}
\usepackage{bm}
\usepackage{algorithm}
\usepackage{algorithmic}
\usepackage{pgfplots}
\usepackage{filecontents}
\usepackage{subcaption}
\usepackage{wrapfig}

\newcommand{\holthen}{{\texttt{\color{cyan}>>}}}
\newcommand{\holthenone}{{\texttt{\color{cyan}>-}}}
\newcommand{\holquoted}[1]{{\color{brown}\textrm{`}#1\textrm{'}}}
\newcommand{\holdoublequoted}[1]{{\color{brown}\textrm{``}#1\textrm{''}}}
\usepackage{alltt}

\newcommand{\ITP}{\textsc{itp}}
\newcommand{\MDP}{\textsc{mdp}}
\newcommand{\ATP}{\textsc{atp}}
\newcommand{\HOL}{\textsc{Hol4}}
\newcommand{\COQ}{\textsc{Coq}}
\newcommand{\HOLLIGHT}{\textsc{Hol Light}}
\newcommand{\METAMATH}{\textsc{metamath}}
\newcommand{\LEAN}{\textsc{Lean}}

\newcommand{\ie}{\textit{i.e.}}

\newcommand{\sref}[2]{\hyperref[#2]{#1~\ref{#2}}} 

\usepackage{amsmath}
\DeclareMathOperator*{\softmax}{Softmax}
\definecolor{azure(colorwheel)}{rgb}{0.0, 0.5, 1.0}
\definecolor{awesome}{rgb}{1.0, 0.13, 0.32}
\usepackage{hyperref}

\title{TacticZero: Learning to Prove Theorems from Scratch with Deep Reinforcement Learning}

\author{%
  Minchao Wu \\
  Research School of Computer Science\\
  Australian National University\\
  Canberra, ACT, Australia \\
  \texttt{Minchao.Wu@anu.edu.au} \\
  \And
  Michael Norrish \\
  Data61, CSIRO \\
  Canberra, ACT, Australia \\
  \texttt{Michael.Norrish@data61.csiro.au} \\
  \AND
  Christian Walder \\
  Data61, CSIRO \\
  Canberra, ACT, Australia \\
  \texttt{Christian.Walder@data61.csiro.au} \\
  \And
  Amir Dezfouli \\
  Data61, CSIRO \\
  Canberra, ACT, Australia \\
  \texttt{Amir.Dezfouli@data61.csiro.au} \\
}

\begin{document}

\maketitle

\begin{abstract}
We propose a novel approach to interactive theorem-proving (\ITP{}) using deep reinforcement learning. The proposed framework is able to learn proof search strategies as well as tactic and arguments prediction in an end-to-end manner. We formulate the process of \ITP{} as a Markov decision process (\MDP{}) in which each state represents a set of potential derivation paths. This structure allows us to introduce a novel backtracking mechanism which enables the agent to efficiently discard (predicted) dead-end derivations and restart from promising alternatives. We implement the framework in the \HOL{} theorem prover. Experimental results show that the framework outperforms existing automated theorem provers (\textit{i.e.} \textit{hammers}) available in \HOL{} when evaluated on unseen problems. We further elaborate the role of key components of the framework using ablation studies.

\end{abstract}


\section{Introduction}
Interactive theorem-proving~(\ITP{}) is the process of humans interacting with a computer system to develop formal proofs of mathematical theorems.
In \ITP{}, a human user can define mathematical objects, and then guide the computer to prove theorems about them using commands called \emph{tactics}. Tactics are programs that embody high-level proof strategies such as simplification and induction.
Successful tactic application converts the current target \emph{goal} (a theorem to be proved) into zero or more subgoals that remain to be proved.

Proofs written in an \ITP{} system are ultimately checked by a machine, and therefore, they are much more trustworthy than pencil-and-paper proofs.
For this reason, \ITP{} has gained success in both formalizing mathematics, and in verifying computer programs~\citep{DBLP:journals/jar/Leroy09,keplerconj2017}.
Nevertheless, because a great amount of formal detail needs to be spelled out when using an \ITP{} system, substantial human inputs are required to fill the gaps between steps in a proof.
In particular, in terms of tactic-based theorem proving, human guidance in the selection of both tactics, \emph{and} the arguments to tactics is crucial to a successful proof.
This requires expert knowledge both of the relevant mathematics, and the particular \ITP{} system being used.
This requirement for expertise has in turn limited the application of \ITP{} systems.

One promising line of work
to address this limitation has been to use machine learning methods to replace the role of human experts in \ITP{}. Existing learning-based approaches have two noticeable characteristics: learning from human examples for tactic prediction and using a fixed proof search strategy (\textit{e.g.} breadth/best-first search (BFS)) that is separate from the learning process to find proofs. The dependency on human examples limits the application of this kind of techniques depending on the availability and quality of the human proofs in the domain that the proof is required. There are potentially infinitely many proofs of a theorem, but there are only one or two of them exist in the library without any gurantee of being optimal. For this reason, it is essential for an agent to be able to explore different proofs by itself in order to generalize well and capture the underlying mathematical knowledge. On the other hand, fixed proof search strategies tend to be artificial and suboptimal. Search strategies such as BFS are usually computationally expensive in time (due to excessive I/O operations needed to communicate with the external theorem prover) and space (due to needless proof states resulting from redundant expansion of previous proof states), and are thus not suitable for learning algorithms that require fast iteration.

\begin{figure*}[t]
  \vskip 0.2in
  \begin{center}
  \centerline{\includegraphics[scale=1]{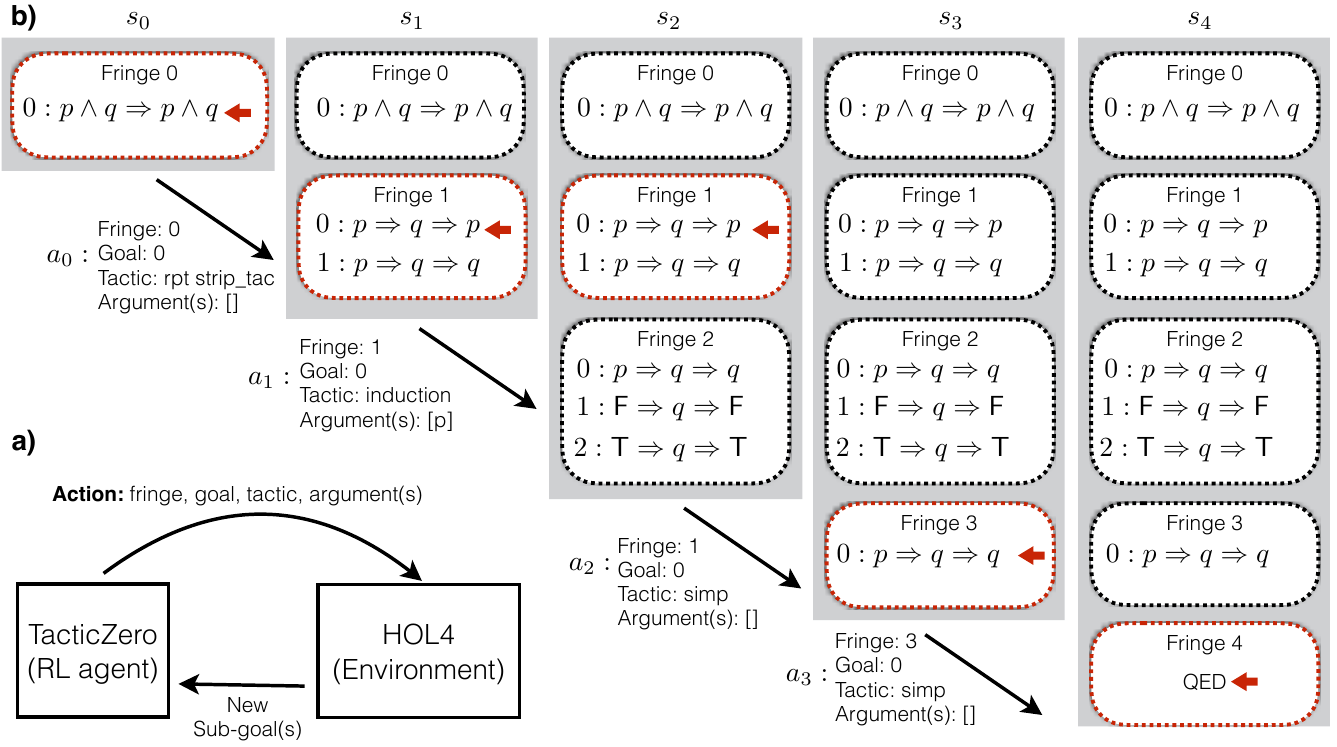}}
  \caption{\textit{a)} Interaction between the RL agent and the environment.
  \textit{b)} An example scenario.
An action ($a_i$) consists of a fringe, a goal, a tactic and its arguments.
Each state consists of multiple fringes and each fringe includes a set of goals that are sufficient to be proved to derive the original target theorem.
At each state, the agent evaluates goals within each fringe and decides which fringe is more promising, and within each fringe which goal to work on first (shown by red arrows).
Given this choice, the agent then selects a tactic and appropriate arguments.
At each state the agent can cease working on the newly added fringe, and backtracks to an older fringe if the new one is predicted to be hard.
This process continues until an empty fringe is derived ($s_4$).
  }
  \label{fig:example}
\end{center}
\vskip -0.2in
\end{figure*}

We propose a reinforcement-learning (RL) framework to learning \ITP{} that addresses the above limitations of the existing approaches.
The RL agent interacts~(\hyperref[fig:example]{Figure~\ref{fig:example}a}) with the \HOL{} interactive theorem prover~\citep{10.1007/978-3-540-71067-7_6} and mimics the behavior of human \ITP{} experts. Given a proof state, a human expert often decides not only what tactic and arguments to apply, but also whether or not to go back to a previous proof state and restart the proof from there. This kind of situation occurs frequently in \ITP{} as careless tactic applications can quickly result in unmanageable proof states. An experienced \ITP{} expert would have the intuition on when and where to backtrack. Our agent learns proof search strategies in a similar manner by learning to choose promising derivations and subgoals to attack, as well as the tactics and their arguments to apply to the goals, without using the limited examples in the library.

\textbf{Our Contributions} are as follows.
\begin{itemize}
\item We provide a formulation of \ITP{} as a Markov decision process (\MDP{})~\citep{Bel} in which flexible state representations enable tracking multiple subproofs during the search process. The structure of the action space allows for tactics to have both theorem names and terms as input arguments.

\item We propose an RL architecture using multiple recurrent and feed-forward neural network modules to solve the above \MDP{}.
We use policy gradients~\citep{10.1007/BF00992696} to jointly learn to apply actions at the backtracking, goal, tactic and argument levels.

\item We implement the framework in the \HOL{} theorem prover and show that it outperforms state-of-the-art automated theorem provers (\textit{i.e.} \textit{hammers}~\citep{10.1145/2676724.2693173}) available in \HOL{} including E~\citep{SCV:CADE-2019}, Z3~\citep{10.1007/978-3-540-78800-3_24} and Vampire~\citep{10.1007/978-3-642-39799-8_1}. We further use ablation studies to establish the contribution of key components of our architecture in achieving the overall performance.
\end{itemize}

\section{Interactive Theorem Proving with Reinforcement Learning}
\begin{wrapfigure}{R}{5cm}
\begin{alltt}
\small\textbf{Theorem} example\textbf{:}
{\color{brown} \(p \land q \;\Rightarrow p \land q\)}
\textbf{Proof}
  rpt strip_tac
  \holthenone{} \textit{\color{orange}(* Induct_on \holquoted{p} *)} simp[]
  \holthenone{} simp[]
\textbf{QED}
\end{alltt}
\caption{A Simple \HOL{} Proof.
 }
\label{fig:simple-hol-proof}
\end{wrapfigure}

We begin with an overview of interactive proving in \HOL{} by a simple example (corresponding to the script in \autoref{fig:simple-hol-proof}).
Suppose we want to prove the theorem $p \wedge q \Rightarrow p \wedge q$ where $p$ and $q$ are propositional variables (\ie, boolean).
This expression is called our initial \textit{goal}.
Since it is an implication, we will assume the antecedent, and then use that to prove the conclusion.
This can be done by applying a \textit{tactic}, \texttt{rpt~strip\_tac} to the goal.
This tactic does not take any arguments.

\HOL{} tells us that the tactic application results in two new \textit{subgoals}.
In this case, they are $p \Rightarrow q \Rightarrow p$ and $p \Rightarrow q \Rightarrow q$.%
\footnote{\HOL{} actually moves $p$ and $q$ to what it calls the goal's \emph{assumption-list}; for simplicity here, we represent that list with chained implications in one formula.}
At this stage, both subgoals need to be proved in order for the proof to succeed.
Suppose we choose to first prove $p \Rightarrow q \Rightarrow p$.
We notice that because $p$ and $q$ are boolean variables, it should be possible to prove the goal by cases.
This can be done by applying the \texttt{Induct\_on}  tactic to the goal.
The \texttt{Induct\_on} tactic takes a single \textit{term} argument, that is,  an arbitrary \HOL{} expression.
In this case, it will be the variable $p$.

\newcommand{\holindonp}{\texttt{Induct\_on\,\textrm{`}p\textrm{'}}}
After applying \holindonp{} to $p \Rightarrow q \Rightarrow p$, two more subgoals are generated, with the variable $p$ replaced by truth values $\mathsf{T}$ and $\mathsf{F}$ respectively.
At this stage, we have three subgoals to deal with.
One is the earlier goal that we chose to defer: $p \Rightarrow q \Rightarrow q$.
The other two are the newly generated $\mathsf{T} \Rightarrow q \Rightarrow \mathsf{T}$ and $\mathsf{F} \Rightarrow q \Rightarrow \mathsf{F}$.

One might decide at this stage that the previous tactic application of \holindonp{} has made the situation worse because it has left us with more goals to prove.
A human expert may now look back at the remaining subgoals before the application of \holindonp{} and try another tactic.
Fortunately, there is a tactic called \texttt{simp} that can prove the goal $p \Rightarrow q \Rightarrow p$ directly.
The \texttt{simp} tactic takes a list of theorem names as its argument.
In this case, an empty argument list is sufficient to prove the goal. Similarly, \texttt{simp[]} proves the remaining $p \Rightarrow q \Rightarrow q$ goal as well.

\subsection{Modeling \ITP{} with \MDP{}}

\textbf{States} A proof attempt starts with a main goal $g\in \mathbb{G}$.
At any point in a proof attempt, there is a set of goals that all need to be proved. We refer to these finite sets of goals as \textit{fringes}. In the context of our framing as an \MDP{}, multiple fringes will be generated and maintained in such a way that the main goal will be proved if everything in any one fringe is proved. Thus a fringe represents a particular path along which we may continue in an attempt to complete a single valid \HOL{} proof of the main goal. In contrast, always choosing the most recently generated fringe would be equivalent to never \textit{backtracking} in \HOL. The \MDP{} state $s$ is therefore a finite sequence of fringes
and we denote the $i$-th fringe in state $s$ by $s(i)$. See \hyperref[fig:example]{Figure~\ref{fig:example}b} for an example state sequence.

\textbf{Actions} An action is a 4-tuple $(i, j, t, \bm{c}) \in \mathbb{N}\times\mathbb{N}\times \mathcal{T}\times\mathcal{A}^*$.
Intuitively, given a state $s$, $i$ selects the $i$-th fringe in $s$ and $j$ selects the $j$-th goal within fringe $s(i)$.
Then, $t$ is the \HOL{} tactic that we select from the set $\mathcal T$ of possible \HOL{} tactics, and $\bm{c}$ is the possibly empty list of arguments that accompany $t$ (see \hyperref[fig:example]{Figure~\ref{fig:example}b}).
An argument in $\mathcal{A}$ is either a theorem name, or a \HOL{} term.
The existence of arguments makes the action space arbitrarily large---terms may be arbitrarily defined, and there are thousands of theorems that can be given as arguments to the tactic.
This necessitates an RL algorithm that can handle large action and state spaces, for which we select policy gradients~\citep{10.1007/BF00992696}.

\textbf{Rewards} Feedback is received from \HOL{} after each tactic application. If successful, it generates a set of new (sub-)goals such that proving all of them proves $g$; if this set is empty, then $g$ is itself proven. A tactic application may fail, either with an error indicating that it is not applicable, or with a timeout caused by \HOL{} exceeding a maximum computation time which we set. In terms of our framing as an \MDP{}, different numeric rewards are associated with the different cases described above.

\textbf{\MDP{} State Transition} A proof attempt always starts with a single main goal $g$, and so the initial state is a single fringe containing one element, $g$.
Given a state $s$, performing an action $(i,j,t,\bm{c})$ may not change the state.
This happens when the tactic times out, has no effect, or is not applicable.
Otherwise, the application of the tactic generates a set of new subgoals. This set $G$ of new subgoals may be empty, indicating that goal $s(i)(j)$ is immediately proved by tactic $t$.
In any case, a new fringe is then constructed by first copying fringe $s(i)$, and then replacing the goal $s(i)(j)$ with the new subgoals $G$.
Then we construct a new state $s'$ by adding the new fringe to state $s$. Each state constructed in this way maintains all possible partial proof attempts. The size of each state is \textit{linear} in the number of timesteps so that the manipulation and storage of states are efficient.

\textbf{Termination} The \MDP{} terminates when an empty fringe is constructed, which implies that we can recover a proof of the main goal. In this case, the agent receives a positive reward. The process is also terminated if the timestep budget is exceeded, in which case the proof attempt is unsuccessful and the agent receives a negative reward. The state $s_4$ in \hyperref[fig:example]{Figure~\ref{fig:example}b} is an example of a terminal state.

\begin{figure}[t]
  \vskip 0.2in
  \begin{center}
  \centerline{\includegraphics[scale=1.1]{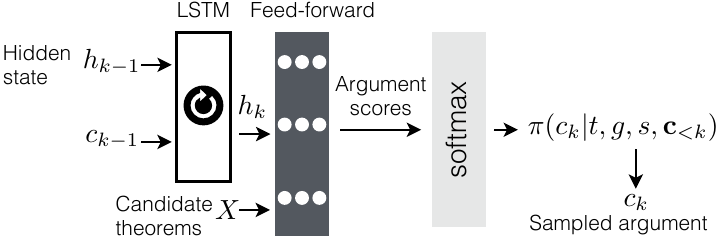}}
  \caption{The argument network. The aim is to select arguments from candidates theorems $X$ for a
  tactic $t$ to be applied on goal $g$. To achieve this, the hidden state of an LSTM neural network are initialized using $g$. The LSTM
  layer takes the previously chosen argument as an input ($c_{k-1}$) and through a feed-forward and a softmax layer generates the probability of selecting
  each candidate theorem for the next argument ($c_k$). $c_0$ is initialized to $t$.}
  \label{fig:arg_net}
\end{center}
\vskip -0.2in
\end{figure}

\subsection{A solution to the MDP formulation}
\label{RLAA}

\textbf{Encoding \HOL{} expressions} Each \HOL{} expression is represented by a 256-dimensional vector, which is learned by a transformer based autoencoder~\citep{kingma2014autoencoding} for sequences pre-trained on 4360 definitions and theorems (tokenized and represented in Polish notation) in the \HOL{} library. For implementation, we use a customized version of the pytorch-seq2seq\footnote{\url{https://github.com/IBM/pytorch-seq2seq}}.

\textbf{Selecting fringes} Let $\mathbb{G}\subseteq\mathbb{R}^{256}$ be the space of goals; we learn a function 
\begin{align}
V_{\mathrm{goal}}: \mathbb{G} \rightarrow \mathbb{R},
\end{align}
which, intuitively, represents how easy a goal is to prove. Given a state $s$ we define the score $v_{s(i)}$ of its $i$-th fringe as 
\begin{align}
v_{s(i)} = \Sigma_{g\in s(i)}V_\mathrm{goal}(g). 
\end{align}
Summing in this way reflects a simplifying design choice; namely that \textit{1)} to prove $s(i)$ we must prove all its consituent goals, \textit{2)} that $V_{\mathrm{goal}}$ outputs logits, and \textit{3)} as we sum logits that the probability of solving each goal is independent given its numeric representation.

To choose a fringe to work on, the agent samples from the discrete distribution with probabilities
\begin{align}
\pi_\mathrm{fringe}(s) = \softmax(v_{s(0)},...,v_{s(|s|-1)}).
\label{eqn:pifringe}
\end{align}
After selecting a fringe, by default we select the first goal in that fringe to work on, because all of the goals within a fringe have to be proved in order to prove the main goal, and the order in which they are proven is irrelevant.

\begin{figure*}[t]
  \vskip 0.2in
  \begin{center}
  \centerline{\includegraphics[scale=1]{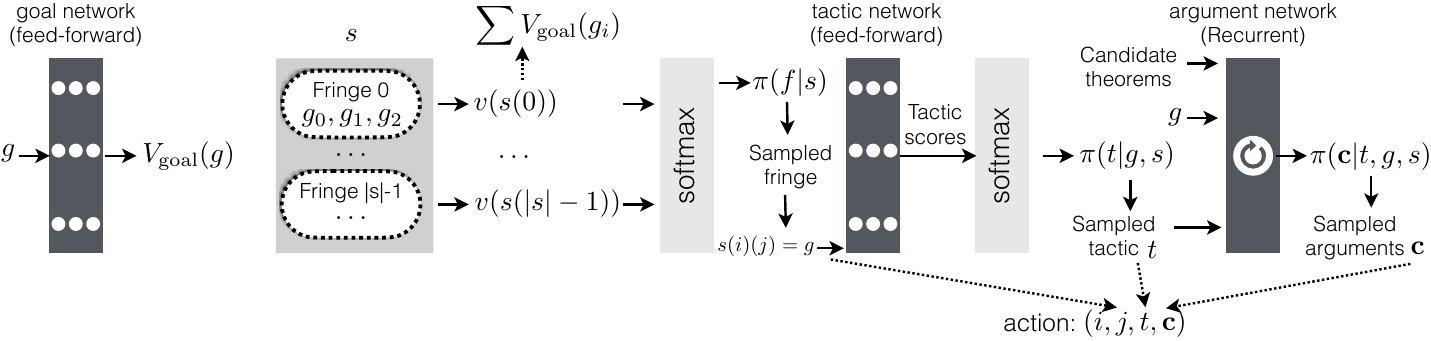}}
  \caption{Structure of the model. The functionality of each network is described in~\autoref{RLAA}.
  }
   \label{fig:model}
\end{center}
\vskip -0.2in
\end{figure*}

\textbf{Selecting tactics} Suppose that we have selected a goal $g$ to work on. We learn a function
\begin{align}
 	V_\mathrm{tactic}: \mathbb{G}\rightarrow \mathbb{R}^D,
 \end{align}
 where $D$ is the total number of tactics allowed, which estimates the effectiveness of each tactic for proving $g$. To select the tactic, we again sample from the discrete distribution, here with probabilities
 \begin{align}
   \pi_\mathrm{tactic}(g) = \softmax(V_\mathrm{tactic}(g)).
   \label{eqn:pitactic}   
\end{align}

 \textbf{Selecting a list of arguments} Suppose that we have selected a goal $g$ and a tactic $t$. If $t$ takes arguments, we need to select a list of them. This process is naturally modeled by a recurrent neural network architecture as shown in \autoref{fig:arg_net}.

 The argument policy recurrence takes three pieces of data as its input. The first is a set of candidate theorems (or a set of candidate terms if $t$ takes a term) that the policy can choose from. The candidate theorems are usually the theorems proved earlier than $g$ in the library, and the candidate terms are the variables that occur in $g$. The second is the previously selected argument, which is initialized to the selected tactic. The third is a latent state variable which carries the sequential information of previously selected arguments over the argument generation process. This is because the arguments to tactics like \texttt{simp} generally depend on each other.

 The first output of the recurrent module is a set of scores associated with the candidate arguments. We softmax these scores and sample to get one argument.
An LSTM also produces a second output which is the latent state variable that carries the sequential information. The recurrent process is terminated when it reaches a pre-defined maximum length $L$ which we set for the argument list.
\textbf{Computing policy gradients} Given a state $s$, the selection process described above now allows us to define our policy distribution in terms of the following factors.
\begin{itemize}
\item $\pi(f|s)$, the probability of selecting fringe $f$ given $s$ induced by $\pi_\mathrm{fringe}$ of \autoref{eqn:pifringe}. Recall that, subsequently, the first goal $g$ in $f$ is always selected.
\item $\pi(t|s,g)$, the probability of selecting tactic $t$ given $s$ and $g$, induced by $\pi_\mathrm{tactic}$ of \autoref{eqn:pitactic}.
\item $\pi(c_l|s,g,t,\bm{c}_{<l}); l = 1, 2, \dots, L$, the probability of selecting the $l$-th element of the argument list $\bm{c}$ given $s, g$, $t$ and the previously generated elements of the argument list. This is induced by the recurrent argument selection, where $L$ is the fixed length of arguments, and $\bm{c}_{<l} = (c_1,...,c_{l-1})$.
\end{itemize}

Denoting the action by $a$, the policy therefore factorises as
\begin{align}
	\pi_{\theta}(a|s) = \pi(f|s)\pi(t|s,g) \prod_{l=1}^{L} \pi(c_{l}|s,g,t,\bm{c}_{<l}),
\end{align}
where $\theta$ represents the parameters of $\{V_\mathrm{goal},V_\mathrm{tactic},V_\mathrm{arg}\}$. Denoting the $m$-th reward by $r_m$, and the \MDP{} trajectory by $\tau=(s_0, a_0, r_0, s_1, a_1, r_1, \dots, s_M, a_M, r_M)$, our objective function is the usual sum of discounted expected rewards with discount factor $\gamma\in(0,1]$, \ie,

\begin{align}
	J(\theta)=\smash[t]{\mathbb E_{\tau\sim \pi_\theta}\Big[\sum_{m=0}^M \gamma^m r_m\Big]}.
\end{align}
Despite the large action space at hand, optimisation in $\theta$ by gradient descent is made tractable  by substituting the classic REINFORCE estimator~\citep{10.1007/BF00992696} for $\nabla_\theta J$.


\section{Experiments}
\subsection{Learning}
\label{sec:experiments}
\textbf{ITP settings}
The learning task is to train an agent that is capable of proving theorems using a given set of tactics in \HOL{}'s core library.
The tactics allowed to be used by the agent are listed in \autoref{tab:tacticset}.
We collect 1342 theorems that are known to be provable using the given set of tactics in \HOL{}'s core library and randomly split them into training and testing sets in an 80:20 ratio.
The training data merely provides the agent with the statement of theorems for it to learn to prove by itself. We set the timestep budget to be 50, and the execution time limit of each tactic application to be 0.1 seconds. Given a goal $g$, the candidate set of theorems that can be chosen as arguments to a tactic is all the theorems which come from the theories mentioned in $g$ and occur before $g$ in the library.

\begin{table}[ht]
\caption{Tactics that can be used by the agent. \\~
}
\label{tab:tacticset}
\centering
\begin{tabular}{ll}
\toprule
 Tactics &   Argument types \\
    \midrule
\texttt{drule}, \texttt{irule} & single theorem \\
\texttt{fs}, \texttt{metis\_tac}, \texttt{rw}, \texttt{simp} & list of theorems \\
\texttt{Induct\_on} & single term  \\
\texttt{eq\_tac}, \texttt{strip\_tac}  &    none \\
\bottomrule
\end{tabular}
\end{table}

\textbf{Reward shaping}
If a theorem is proved, a positive terminal reward depending on the difficulty of the target theorem is given. If the rate at which the theorem is being proved in earlier rollouts is above average, then the reward is 5. Otherwise, the reward is 15. If the proof attempt fails, the agent receives a terminal reward of -5.
We also encourage the agent to ``make progress'' by giving it a 0.1 reward when new subgoals are generated and a 0.2 reward when a subgoal is solved. For all other outcomes after each timestep, a reward -0.1 is given. The negative terminal reward reduces the tendency of the agent to maximize the reward by making irrelevant progress that does not lead to a successful proof of the main goal. By nature of the tactics in~\autoref{tab:tacticset}, repeatedly applying the same tactic to the same goal will eventually result in unchanged fringes, in which case the -0.1 reward will be given.

\textbf{Replaying}
One difficulty during training is that positive rewards are sparse. The agent may find a proof of a difficult theorem by accident, but then take many episodes to prove it again.
To help the agent recall its successes, we maintain a replay buffer of earlier successful proofs of each theorem.
During training, if the agent fails to prove a theorem it was previously able to prove, replaying will be triggered so that the agent is walked through one of the 5 most recent successful proof and parameters updated correspondingly. This represents a departure from pure policy gradients, but worked well in our experiments, presumably because the update remains in the high dimensional direction $\nabla_\theta J$ and therefore differs from a precise \textit{on-policy} update only by the ad-hoc choice of learning rate.

\textbf{Training} We jointly train the policies using RMSProp~\citep{tieleman2012lecture} with a learning rate of $5\times 10^{-5}$ for each policy. The discount factor $\gamma$ is set to be $0.99$. The structure of our model is illustrated in~\autoref{fig:model}. It takes two weeks with a single Tesla P100-SXM2 GPU and a single Intel(R) Xeon(R) CPU E5-2690 v4 @ 2.60GHz to complete 800 iterations.
\subsection{Evaluation}
In this section, we study the theorem proving capability of TacticZero by comparing it with three state-of-the-art automated theorem provers (Z3, E and Vampire) available as hammers in \HOL{}.
When choosing arguments (premises), TacticZero and each hammer will use the same set of candidate theorems as described in~\autoref{sec:experiments}.
For each hammer, we first call them with the default evaluation mode, in which case the provers will try to use all the theorems in $C$ to prove the target theorem. We then call the provers by using holyHammer's learning-based mechanism~\citep{10.1145/2676724.2693173} of premise selection, which will first choose $n$ theorems from $C$ and then send them to the provers as premises. We also provide four additional baselines: an untrained TacticZero agent, breadth first search (BFS) and depth first search (DFS) with random choice of tactics and arguments, and always calling a single \texttt{metis\_tac[]} for each theorem.

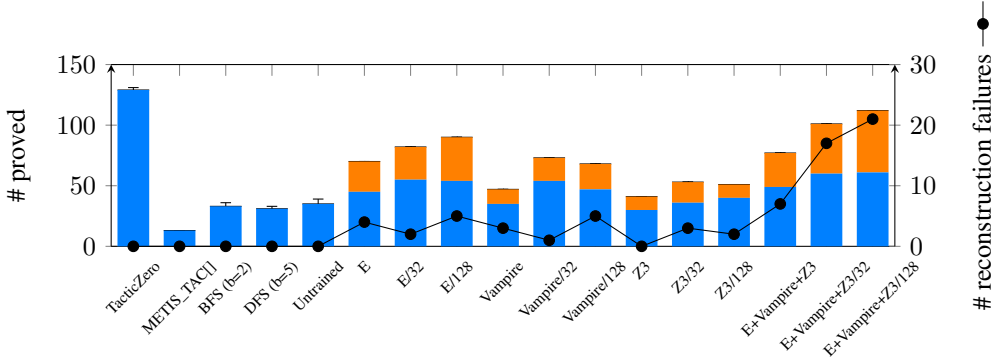
\begin{figure}[ht]
\centering
\begin{tikzpicture}
  \begin{axis}[
    axis y line=left,
    width=12cm,
    height=4cm,
    ybar stacked,
    bar width=12,
    ymin=0,
    ymax=150,
    xticklabel style={rotate=45,font=\tiny},
    xticklabels from table={data.dat}{n}, 
    xtick=data, 
    enlarge x limits=0.03, 
    ylabel=\# proved
  ]
    \addplot[draw=none,fill=azure(colorwheel)] table[x expr=\coordindex,y=ol] {data.dat};
    \addplot[draw=none,fill=orange] table[x expr=\coordindex,y=st] {data.dat};
\addplot[
        error bars/.cd,
            y dir=both,
            y explicit
    ] table[x expr=\coordindex,y=per,y error=err] {data.dat};
  \end{axis}
\begin{axis}[
    width=12cm,
    height=4cm,
    ymin=0,
    ymax=30,
axis y line=right,
ylabel style = {align=center},
ylabel={\# reconstruction failures~\ref{recon-failures}},
axis x line=none,
    xticklabel style={rotate=45,font=\tiny},
    xticklabels from table={data.dat}{n}, 
    xtick=data, 
    enlarge x limits=0.03,
  ]
  \addplot[color=black,mark=*] table[x expr=\coordindex,y=rf] {data.dat};
  \label{recon-failures}
\end{axis}
\end{tikzpicture}
\caption{Performance of TacticZero compared to hammers and additional baselines. E (Z3, Vampire resp.) indicates the performance of the E prover when using all available theorems as premises. E/32 indicates the performance of E prover when using premise selection to choose 32 theorems as premises.  Parameter $b$ in BFS and DFS is the branching factor that controls the number of expansions for each node. The reconstruction of proofs found by hammers is not always successful. The black line plot shows the number of failed reconstructions of each method (note different scale on right). The blue portion of each bar shows the number of theorems that are also proved by TacticZero.}
\label{fig:baselines}
\end{figure}

\autoref{fig:baselines} shows the number of theorems in the testing set proved by each method given a timeout of 10 seconds. \texttt{metis\_tac} as the simplest baseline can prove only 13 (out of 268) testing problems, suggesting that the majority of the testing problems are non-trivial. TacticZero solves the largest number of theorems, which is 132. In contrast, E, Vampire and Z3 solve 68, 47 and 41 theorems rescpectively, when using all the available facts as premises. The performance of hammers improves when learning-based premise selection is enabled (see E/128 for example). TacticZero is able to prove 18\% more theorems (132 as opposed to 112 given by E+Vampire+Z3/128) even when we combine all the hammers together with premise selection enabled.

There is also an overlap between the theorems proved by TacticZero and those proved by other methods. For example, there are 61 theorems proved by both TacticZero and E+Vampire+Z3/128, meaning that TacticZero proved 71 theorems not found by combining E, Vampire and Z3, and the hammers working together proved 51 theorems not found by TacticZero. It is thus possible to achieve better performance by combining TacticZero and hammers in practice. We also note that learnt proofs can be dramatic improvements on what human authors wrote:~\hyperref[fig:tacticzero-out]{Figure~\ref{fig:tacticzero-out}a} shows an example of this.

The BFS/DFS baselines have the second worst performance among all the baselines. This suggests that proper selection of tactics and arguments is essential even when using a principled proof search strategy. In the next section, we further study the impact of different proof search strategies when proper tactic and argument policies are available.

\subsection{Ablation study}
We now study the proof search strategy learned by TacticZero by comparing it with other fixed search strategies including depth-first search, breadth/best-first search with different branching factors, and greedily expanding the latest fringe. Learned tactic and argument policies are available for all the agents in this experiment. All the agents are given the same amount of timestep budget which is 50.

\begin{table}
  \caption{Ablation study for different proof search strategies in comparison with the full TacticZero. Stochastic means that when branching, the agent follows the tactic policy stochastically by sampling from the policy distribution; topk means that the agent chooses deterministically best $k$ tactics suggested by the tactic policy, where $k=b$.
\\~}
  \label{tab:ablation}
  \centering
\begin{tabular}{lccc}
\toprule
 Methods &   \# proved & mean \# timesteps & mean length of proof \\
    \midrule
\texttt{BFS/stochastic~(b=2)} & 99 & 15.32 & 3.25 \\
\texttt{BFS/stochastic~(b=3)} & 79 & 17.01 & 2.60 \\
\texttt{BFS/topk~(b=2)} & 116 & 15.85 & 3.10 \\
\texttt{BFS/topk~(b=3)} & 92 & 14.79 & 2.36 \\
\texttt{DFS/stochastic~(b=2)} & 83 & 8.33 & 6.00 \\
\texttt{DFS/stochastic~(b=3)} & 96 & 9.71 & 6.21 \\
\texttt{DFS/topk~(b=2)} & 86 & 8.04 & 5.86 \\
\texttt{DFS/topk~(b=3)} & 90 & 9.78 & 6.36 \\
\texttt{Latest~fringe} & 112 & 11.07 & 6.94 \\
\midrule
\texttt{TacticZero} & 132 & 11.72 & 5.00 \\
\bottomrule
\end{tabular}
\end{table}

The results of the experiment are shown in \autoref{tab:ablation}. The largest number of proved theorems is given by the search strategy learned by TacticZero. The BFS family tends to find shorter proofs but generally takes more timesteps to find a proof. This is because the depth of BFS is determined by the timestep budget and the branching factor\footnote{For example, with a timestep budget = 50 and a branching factor = 2, the maximal depth of the search tree would be 5, which means that the maximal length of proofs that can be found by such an agent is 5.}, and there might be redundant expansion in a single level. On the other hand, the DFS family tends to find proofs quickly (as indicated by the timesteps column and row 5-8)
but the proofs are rather long. This is because DFS lacks the ability to backtrack further than 1-level up in the search tree. This prevents the agent finding alternative and potentially shorter derivations by restarting the proof at a higher level when it is stuck with a particular node. Greedily expanding the latest fringe behaves similarly to DFS, but instead of backtracking, it stays with the same node forever and queries the tactic and argument policies for alternatives until the node becomes expandable. This strategy is slightly better than the DFS family in terms of the number of provable theorems, but it takes more timesteps to find proofs, and tends to find longer proofs than those discovered by DFS.

\begin{figure}[htb]
    \begin{minipage}[hb]{.45\textwidth}
\begin{alltt}
\tiny
\textbf{Theorem} EVERY2_DROP\textbf{:}
  {\color{brown}\(\forall\)R l1 l2 n. EVERY2 R l1 l2 \(\Rightarrow\)
    EVERY2 R (DROP n l1) (DROP n l2)}
\textbf{Proof}
  Induct_on \holquoted{n}
  \holthenone{} (strip_tac \holthen{} fs[])
  \holthenone{} (Induct_on \holquoted{l1} \holthenone{} fs[]
      \holthenone{} (rpt strip_tac \holthen{} fs[]))
\textbf{QED}
\textit{(*}
  \textit{\color{orange}(* Original human proof *)}
  rpt strip_tac \holthen{} IMP_RES_TAC LIST_REL_LENGTH
  \holthen{} Q.PAT_ASSUM \holquoted{LIST_REL P xs ys} MP_TAC
  \holthen{} ONCE_REWRITE_TAC [GSYM TAKE_DROP] \holthen{} rpt strip_tac
  \holthen{} ONCE_REWRITE_TAC [TAKE_DROP] \holthen{} Cases_on \holquoted{n \(\le\) LENGTH l1}
  \holthenone{} metis_tac [EVERY2_APPEND,LENGTH_DROP,LENGTH_TAKE]
  \holthen{} fs [GSYM NOT_LESS] \holthen{} \holquoted{LENGTH l1 \(\le\) n} by numLib.DECIDE_TAC
  \holthen{} fs [DROP_LENGTH_TOO_LONG] \holthen{} rfs [DROP_LENGTH_TOO_LONG]
\textit{*)}
\end{alltt}
        \subcaption{A proof discovered by TacticZero.
The proof has been minimized through the removal of redundant arguments provided to the \texttt{fs} tactic.
The human proof of the same theorem has significantly more steps (notwithstanding its access to a much bigger pool of basic tactics).
}\label{fig:tacticzero-proof}
    \end{minipage}
    \hfill
    \begin{minipage}[hb]{.48\textwidth}
        \includegraphics[width=\textwidth]{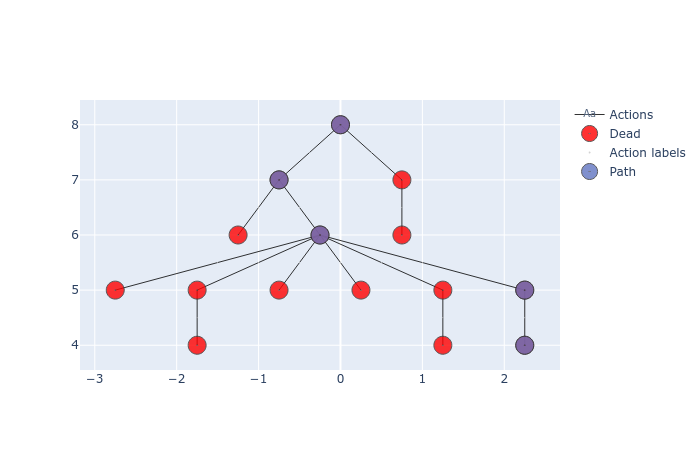}
        \subcaption{A proof search of the theorem $\forall\;x\;s.\; x \in s \Rightarrow \forall f.\; f(x) \in \texttt{IMAGE}\;f\;s$.  Red nodes represent the fringes that never lead to a successful proof, and blue nodes consist of a path from which a proof can be reconstructed.
This proof search is neither a BFS nor DFS, but a unique strategy involving backtracking.
}\label{fig:2}
    \end{minipage}
    \label{fig:tacticzero-search}
    \caption{Example proofs and proof search performed by TacticZero.}
\label{fig:tacticzero-out}
\end{figure}

We also note that the learned proof search strategy not only finds the largest number of proofs, but also sits in the middle of the chart in terms of timesteps and length of proof --- it takes fewer timesteps than a DFS agent to find a proof, and finds ``deeper'' proofs that may not be discoverable by a BFS agent within the same timestep budget, and tends to find proofs that are shorter than those found by a DFS agent. In fact, the learned strategy is often neither BFS nor DFS. See~\hyperref[fig:tacticzero-out]{Figure~\ref{fig:tacticzero-out}b} for an example.

\section{Related Work}
\textbf{Automated theorem proving (\ATP)}
The problem setting of machine learning for \ATP{} is rather different from that of \ITP{}. In \ATP{}, one usually works with a ``backbone'' algorithm such as a resolution-based or connection-based algorithm~\citep{letz1994controlled}, upon which machine learning approaches~\citep{10.1007/978-3-662-48899-7_7,NIPS2016_fe9fc289,DBLP:conf/nips/WangTWD17,NIPS2017_b2ab0019,DBLP:journals/corr/LoosISK17,DBLP:journals/corr/abs-1802-03685,NEURIPS2018_55acf853,DBLP:journals/corr/abs-1911-02065,DBLP:journals/corr/abs-1905-13100,DBLP:conf/iclr/LedermanRSL20,DBLP:journals/corr/abs-2103-03798} are built to guide proof search. \ITP{} systems do not use such a ``backbone'' algorithm as their main framework for theorem proving. For this reason, the action space of \ATP{} is different from that of \ITP{}, and machine learning approaches developed for \ATP{} are not compatible with \ITP{}. One the other hand, proofs in \ATP{} are often represented in a low-level language and hard to interpret as high-level mathematical concepts, in contrast to \ITP{} which uses tactics and embodies more human-like mathematical reasoning.

There are also approaches that work by interfacing \ATP{} with \ITP{} systems. These tools are called \textit{hammers}~\citep{DBLP:conf/lpar/PaulsonB10,Kaliszyk2014,Kaliszyk2015,10.1145/2676724.2693173,czajka2018hammer}. These tools work by translating goals in an \ITP{} system to the languages in \ATP{} systems. If a proof is found by an \ATP{} system, the tool then tries to reconstruct the corresponding proof in the \ITP{} system. This process, however, might fail in the situations that the derived proof cannot be fully translated back to the \ITP{} high-level representations.

\textbf{Interactive theorem proving} Machine learning for \ITP{} are more closely related to our work, and most focus on supervised learning from existing human proofs in the library of an \ITP{} system. \textit{TacticToe}~\citep{Gauthier2020} chooses tactics based on recorded human proofs in the library, and uses a distance-weighted $k$ nearest-neighbour classifier~\citep{5408784} for lemma selection~\citep{PxTP2013:Stronger_Automation_for_Flyspeck}. \textit{TacticToe} does not handle the selection of \HOL{} terms as arguments, which is supported by our framework. \textit{GPT-f}~\citep{polu2020generative,DBLP:journals/corr/abs-2102-06203} learns from proofs in the \METAMATH{}~\citep{Megill97metamath:a} (\LEAN{}~\citep{10.1007/978-3-319-21401-6_26} resp.) library, and uses transformer-based~\citep{NIPS2017_3f5ee243} language models to predict proof steps and tactics. For proof search, they both use BFS as a fixed search strategy. In contrast, our approach learns a dynamic proof search strategy that enables backtracking.
\textit{GamePad}~\citep{DBLP:conf/iclr/HuangDSS19} and \textit{ASTactic}~\citep{pmlr-v97-yang19a} are both learning environments for the \COQ{} theorem prover~\citep{coq} and focus on learning from human proofs.
\textit{GamePad} targets specific sub-tasks of \ITP{} such as the algebraic rewriting problem.
\textit{ASTactic} is more general and comes with a deep learning model that learns from human proofs to generate tactics by expanding abstract syntax trees. A beam search is then implemented as the proof search strategy to find proofs. \textit{TacTok}~\citep{10.1145/3428299} improves \textit{ASTactic} by learning also from partial proof scripts, but follows \textit{ASTactic} by using the same beam search as its search strategy.
\textit{ProverBot9001}~\citep{DBLP:journals/corr/abs-1907-07794} learns from proofs in the \textit{CompCert}~\citep{DBLP:journals/jar/Leroy09} project of \COQ{}, and uses recurrent neural networks to predict arguments for tactics. For proof search, \textit{ProverBot9001} expands their search tree by DFS.

There are also approaches including reinforcement learning components. \textit{HOList}/\textit{DeepHOL}~\citep{bansal2019holist,bansal2020learning} trains a proof guidance model to prove theorems in the \HOLLIGHT{} theorem prover~\citep{10.1007/BFb0031814} by continuously expanding training data.
If a proof is found, it is used to generate additional training data, which is used to update the model used for exploration.
Although the process is referred to as a reinforcement learning loop, it uses pre-engineered scoring for premise selection to help find new proofs, and a BFS strategy to find proofs.
In our framework, the agent learns arguments (premise) selection as well as tactic selection without pre-engineered scoring, and manages proof search by itself, all through deep policy gradients.


\section{Conclusion}
\label{sec:conclusion}
We introduced a reinforcement learning framework to learning interactive theorem proving in \HOL{}. Rather than sequentially searching for a \HOL{} proof, our agent exists within the more abstract context of our Markov decision process environment. This environment supports learning tactic prediction as well as proof search strategies in a principled and effective manner without relying on limited human examples in the libraries. We also hope that the MDP formulation opens the possibility of applying other well-developed RL algorithms to \ITP{}. In the future, we plan to overcome the limitation of our framework by allowing free generation of complex \HOL{} terms as arguments to obtain a more refined action space, and using a principled replaying mechanism to increase sample efficiency, as well as training the agent on individual \HOL{} projects such as the CakeML~\citep{10.1145/2535838.2535841} library.

\section*{Acknowledgments}
  We thank the members of Prague automated reasoning group for helpful discussion on an earlier version of this work. In particular, we would like to thank Josef Urban and Thibault Gauthier for their valuable feedback on the prototype of our agent.

\bibliography{example_paper}

\newpage

\appendix

\section{Example proofs}
The subsequent pages include several example proofs presented in the same format as~\hyperref[fig:tacticzero-out]{Figure~\ref{fig:tacticzero-out}a}.

\begin{figure*}[ht]
\begin{alltt}
\mbox{\textit{\color{orange}(* TacticZero proof *)}}

\textbf{Theorem} EVERY_CONJ\textbf{:}
  {\color{brown}\(\forall\)P Q l. EVERY (\(\lambda\)(x:'a). (P x) \(\wedge\) (Q x)) l = (EVERY P l \(\wedge\) EVERY Q l)}
\textbf{Proof}
  rw[] \holthen{} Induct_on \holquoted{l}
  \holthenone{} (rw[listTheory.EVERY_DEF])
  \holthenone{} (rw[listTheory.EVERY_DEF] \holthen{} metis_tac[])
\textbf{QED}
\textit{(*}
  \textit{\color{orange}(* Original human proof *)}
   NTAC 2 GEN_TAC THEN LIST_INDUCT_TAC THEN
   ASM_REWRITE_TAC [EVERY_DEF] THEN
   CONV_TAC (DEPTH_CONV BETA_CONV) THEN
   REPEAT (STRIP_TAC ORELSE EQ_TAC) THEN
   FIRST_ASSUM ACCEPT_TAC);
\textit{*)}


\mbox{\textit{\color{orange}(* TacticZero proof *)}}

\textbf{Theorem} MONO_EXISTS\textbf{:}
  {\color{brown}(\(\forall\)x. P x \(\Rightarrow\) Q x) \(\Rightarrow\) (EXISTS P l \(\Rightarrow\) EXISTS Q l)}
\textbf{Proof}
  rw[listTheory.EXISTS_MEM] \holthen{} metis_tac[]
\textbf{QED}
\textit{(*}
  \textit{\color{orange}(* Original human proof *)}
  Q.ID_SPEC_TAC \holquoted{l} THEN LIST_INDUCT_TAC THEN
  ASM_SIMP_TAC (srw_ss()) [DISJ_IMP_THM]);
\textit{*)}


\mbox{\textit{\color{orange}(* TacticZero proof *)}}

\textbf{Theorem} FLAT_APPEND\textbf{:}
  {\color{brown}\(\forall\)l1 l2. FLAT (APPEND l1 l2) = APPEND (FLAT l1) (FLAT l2)}
\textbf{Proof}
  strip_tac \holthen{} strip_tac \holthen{}
  Induct_on \holquoted{l1}
  \holthenone{} (rw[listTheory.APPEND, listTheory.FLAT])
  \holthenone{} (fs[listTheory.APPEND]
      \holthen{} fs[listTheory.APPEND_ASSOC, listTheory.FLAT])
\textbf{QED}
\textit{(*}
  \textit{\color{orange}(* Original human proof *)}
  LIST_INDUCT_TAC
  THEN REWRITE_TAC [APPEND, FLAT]
  THEN ASM_REWRITE_TAC [APPEND_ASSOC]);
\textit{*)}

\end{alltt}
\label{fig:tactic-zero-example-proof1}
\end{figure*}

\begin{figure*}[ht]
\begin{alltt}
\mbox{\textit{\color{orange}(* TacticZero proof *)}}

\textbf{Theorem} MEM_MAP_f\textbf{:}
  {\color{brown}\(\forall\)f l a. MEM a l \(\Rightarrow\) MEM (f a) (MAP f l)}
\textbf{Proof}
  strip_tac \holthen{} strip_tac \holthen{} strip_tac
  \holthen{} rw[listTheory.MEM_MAP] \holthen{} (metis_tac[listTheory.MEM_MAP])
\textbf{QED}
\textit{(*}
  \textit{\color{orange}(* Original human proof *)}
  PROVE_TAC[MEM_MAP]
\textit{*)}


\mbox{\textit{\color{orange}(* TacticZero proof *)}}

\textbf{Theorem} REVERSE_11\textbf{:}
  {\color{brown}(\(\forall\)l1 l2:'a list. (REVERSE l1 = REVERSE l2) \(\Leftrightarrow\) (l1 = l2)}
\textbf{Proof}
  strip_tac \holthen{} strip_tac
  \holthen{} metis_tac[listTheory.REVERSE_REVERSE]
\textbf{QED}
\textit{(*}
  \textit{\color{orange}(* Original human proof *)}
   REPEAT GEN_TAC THEN EQ_TAC THEN1
     (DISCH_THEN (MP_TAC o AP_TERM \holdoublequoted{REVERSE : 'a list \(\rightarrow\) 'a list}) THEN
      REWRITE_TAC [REVERSE_REVERSE]) THEN
   STRIP_TAC THEN ASM_REWRITE_TAC []);
\textit{*)}


\mbox{\textit{\color{orange}(* TacticZero proof *)}}

\textbf{Theorem} FILTER_COMM\textbf{:}
  {\color{brown}\(\forall\)f1 f2 l. FILTER f1 (FILTER f2 l) = FILTER f2 (FILTER f1 l)}
\textbf{Proof}
Induct_on \holquoted{l}
\holthenone{} rw[]
\holthenone{} rw[]
\textbf{QED}
\textit{(*}
  \textit{\color{orange}(* Original human proof *)}
  NTAC 2 GEN_TAC
  THEN BasicProvers.Induct
  THEN REWRITE_TAC [FILTER]
  THEN GEN_TAC
  THEN REPEAT COND_CASES_TAC
  THEN ASM_REWRITE_TAC [FILTER]);
\textit{*)}

\end{alltt}
\label{fig:tactic-zero-example-proof2}
\end{figure*}

\begin{figure*}[ht]
\begin{alltt}
\mbox{\textit{\color{orange}(* TacticZero proof *)}}

\textbf{Theorem} ABSORPTION\textbf{:}
  {\color{brown}\(\forall\)x:'a. \(\forall\)s. (x IN s) \(\Leftrightarrow\) (x INSERT s = s)}
\textbf{Proof}
  strip_tac
  \holthen{} rw[pred_setTheory.INSERT_DEF]
  \holthen{} fs[pred_setTheory.GSPEC_ETA, pred_setTheory.INSERT_DEF]
  \holthen{} metis_tac[pred_setTheory.SPECIFICATION]
\textbf{QED}
\textit{(*}
  \textit{\color{orange}(* Original human proof *)}
  REWRITE_TAC [EXTENSION,IN_INSERT] THEN
  REPEAT (STRIP_TAC ORELSE EQ_TAC) THEN
  ASM_REWRITE_TAC [] THEN
  FIRST_ASSUM (fn th => fn g => PURE_ONCE_REWRITE_TAC [SYM(SPEC_ALL th)] g)
  THEN DISJ1_TAC THEN REFL_TAC
\textit{*)}


\mbox{\textit{\color{orange}(* TacticZero proof *)}}

\textbf{Theorem} DISJOINT_INSERT\textbf{:}
  {\color{brown}(\(\forall\)(x:'a) s t. DISJOINT (x INSERT s) t \(\Leftrightarrow\) DISJOINT s t \(\wedge\) x NOTIN t}
\textbf{Proof}
  strip_tac \holthen{} strip_tac \holthen{} strip_tac
  \holthen{} fs[pred_setTheory.IN_INSERT, pred_setTheory.INSERT_DEF, pred_setTheory.IN_DISJOINT]
  \holthen{} metis_tac[]
\textbf{QED}
\textit{(*}
  \textit{\color{orange}(* Original human proof *)}
  REWRITE_TAC [IN_DISJOINT,IN_INSERT] THEN
  CONV_TAC (ONCE_DEPTH_CONV NOT_EXISTS_CONV) THEN
  REWRITE_TAC [DE_MORGAN_THM] THEN
  REPEAT GEN_TAC THEN EQ_TAC THENL
  [let val v = genvar (==`:'a`==)
       val GTAC = X_GEN_TAC v
   in DISCH_THEN (fn th => CONJ_TAC THENL [GTAC,ALL_TAC] THEN MP_TAC th)
      THENL [DISCH_THEN (STRIP_ASSUME_TAC o SPEC v) THEN ASM_REWRITE_TAC [],
             DISCH_THEN (MP_TAC o SPEC (\holdoublequoted{x:'a})) THEN REWRITE_TAC[]]
   end,
   REPEAT STRIP_TAC THEN ASM_CASES_TAC (\holdoublequoted{x':'a = x}) THENL
   [ASM_REWRITE_TAC[], ASM_REWRITE_TAC[]]]
\textit{*)}

\end{alltt}
\label{fig:tactic-zero-example-proof3}
\end{figure*}

\begin{figure*}[ht]
\begin{alltt}
\mbox{\textit{\color{orange}(* TacticZero proof *)}}

\textbf{Theorem} INSERT_INTER\textbf{:}
  {\color{brown}(\(\forall\)x:'a. \(\forall\)s t. (x INSERT s) INTER t = (if x IN t then x INSERT (s INTER t) else s INTER t)}
\textbf{Proof}
strip_tac \holthen{} strip_tac \holthen{}
rw[pred_setTheory.INSERT_DEF, pred_setTheory.SPECIFICATION, pred_setTheory.INTER_DEF]
\holthenone{} (rw[pred_setTheory.GSPEC_ETA] \holthen{} metis_tac[])
\holthenone{} (rw[] \holthen{} (rw[pred_setTheory.GSPEC_ETA] >> metis_tac[]))
\textbf{QED}
\textit{(*}
  \textit{\color{orange}(* Original human proof *)}
  REPEAT GEN_TAC THEN COND_CASES_TAC THEN
  ASM_REWRITE_TAC [EXTENSION,IN_INTER,IN_INSERT] THEN
  GEN_TAC THEN EQ_TAC THENL
  [STRIP_TAC THEN ASM_REWRITE_TAC [],
   STRIP_TAC THEN ASM_REWRITE_TAC [],
   PURE_ONCE_REWRITE_TAC [CONJ_SYM] THEN
   DISCH_THEN (CONJUNCTS_THEN MP_TAC) THEN
   STRIP_TAC THEN ASM_REWRITE_TAC [],
   STRIP_TAC THEN ASM_REWRITE_TAC []]);
\textit{*)}


\mbox{\textit{\color{orange}(* TacticZero proof *)}}

\textbf{Theorem} SET_MINIMUM\textbf{:}
  {\color{brown}(\(\forall\)s:'a \(\rightarrow\) bool. \(\forall\)M. (\(\exists\)x. x IN s) \(\Leftrightarrow\) \(\exists\)x. x IN s \(\wedge\) \(\forall\)y. y IN s \(\Leftrightarrow\) M x <= M y}
\textbf{Proof}
rw[]
\holthen{} fs[boolTheory.IMP_CONG, boolTheory.EQ_TRANS, boolTheory.EQ_IMP_THM]
\holthen{} rw[arithmeticTheory.WOP_measure, boolTheory.COND_ABS]
\holthen{} metis_tac[boolTheory.ONE_ONE_THM]
\textbf{QED}
\textit{(*}
  \textit{\color{orange}(* Original human proof *)}
  REPEAT (STRIP_TAC ORELSE EQ_TAC) THENL
  [IMP_RES_THEN (ASSUME_TAC o ISPEC (\holdoublequoted{M:'a\(\rightarrow\)num})) lemma THEN
   let val th = SET_SPEC_CONV (\holdoublequoted{(n:num) IN {M x | (x:'a) IN s}})
   in IMP_RES_THEN (STRIP_ASSUME_TAC o REWRITE_RULE [th]) NUM_SET_WOP
   end THEN EXISTS_TAC (\holdoublequoted{x':'a}) THEN CONJ_TAC THENL
   [FIRST_ASSUM ACCEPT_TAC,
    FIRST_ASSUM (SUBST_ALL_TAC o SYM) THEN
    REPEAT STRIP_TAC THEN FIRST_ASSUM MATCH_MP_TAC THEN
    EXISTS_TAC (\holdoublequoted{y:'a}) THEN CONJ_TAC THENL
    [REFL_TAC, FIRST_ASSUM ACCEPT_TAC]],
   EXISTS_TAC (\holdoublequoted{x:'a}) THEN FIRST_ASSUM ACCEPT_TAC]
\textit{*)}

\end{alltt}
\label{fig:tactic-zero-example-proof4}
\end{figure*}

\begin{figure*}[ht]
\begin{alltt}
\mbox{\textit{\color{orange}(* TacticZero proof *)}}

\textbf{Theorem} INJ_DELETE\textbf{:}
  {\color{brown}(\(\forall\)f s t. INJ f s t ==> \(\forall\)e. e IN s ==> INJ f (s DELETE e) (t DELETE (f e))}
\textbf{Proof}
strip_tac \holthen{} strip_tac \holthen{} strip_tac
\holthen{} fs[] \holthen{} rw[]
\holthen{} (fs[pred_setTheory.INJ_DEF] \holthen{}
    (strip_tac \holthen{} fs[pred_setTheory.IN_DELETE, boolTheory.IMP_DISJ_THM]
     \holthenone{} (metis_tac[pred_setTheory.IN_APP])
     \holthenone{} (fs[] \holthen{} (fs[] \holthen{} (fs[] \holthen{} (metis_tac[]))))))
\textbf{QED}
\textit{(*}
  \textit{\color{orange}(* Original human proof *)}
  RW_TAC bool_ss [INJ_DEF, DELETE_DEF] THENL
  [\holquoted{~(e = x)} by FULL_SIMP_TAC bool_ss
                 [DIFF_DEF,DIFF_INSERT, DIFF_EMPTY, IN_DELETE] THEN
  FULL_SIMP_TAC bool_ss [DIFF_DEF,DIFF_INSERT, DIFF_EMPTY, IN_DELETE] THEN
  METIS_TAC [],
  METIS_TAC [IN_DIFF]]);
\textit{*)}


\mbox{\textit{\color{orange}(* TacticZero proof *)}}

\textbf{Theorem} IMAGE_SURJ\textbf{:}
  {\color{brown}(\(\forall\)f:'a->'b. \(\forall\)s t. SURJ f s t = ((IMAGE f s) = t)}
\textbf{Proof}
strip_tac \holthen{} strip_tac \holthen{} rw[pred_setTheory.SURJ_DEF]
\holthen{} fs[] \holthen{} fs[pred_setTheory.EXTENSION]
\holthen{} fs[pred_setTheory.SPECIFICATION] \holthen{} fs[]
\holthen{} fs[pred_setTheory.IMAGE_applied]
\holthen{} fs[pred_setTheory.IN_APP, boolTheory.RES_EXISTS_THM]
\holthen{} metis_tac[]
\textbf{QED}
\textit{(*}
  \textit{\color{orange}(* Original human proof *)}
  PURE_REWRITE_TAC [SURJ_DEF,EXTENSION,IN_IMAGE] THEN
  REPEAT GEN_TAC THEN EQ_TAC THENL
  [REPEAT (STRIP_TAC ORELSE EQ_TAC) THENL
   [RES_TAC THEN ASM_REWRITE_TAC [],
    MAP_EVERY PURE_ONCE_REWRITE_TAC [[CONJ_SYM],[EQ_SYM_EQ]] THEN
    FIRST_ASSUM MATCH_MP_TAC THEN FIRST_ASSUM ACCEPT_TAC],
   DISCH_THEN (ASSUME_TAC o CONV_RULE (ONCE_DEPTH_CONV SYM_CONV)) THEN
   ASM_REWRITE_TAC [] THEN REPEAT STRIP_TAC THENL
   [EXISTS_TAC (\holdoublequoted{x:'a}) THEN ASM_REWRITE_TAC [],
    EXISTS_TAC (\holdoublequoted{x':'a}) THEN ASM_REWRITE_TAC []]])
\textit{*)}

\end{alltt}
\label{fig:tactic-zero-example-proof4}
\end{figure*}

\end{document}